\documentclass[sigconf]{acmart}
\renewcommand\footnotetextcopyrightpermission[1]{} % removes footnote with conference information in first column

\AtBeginDocument{%
  \providecommand\BibTeX{{%
    \normalfont B\kern-0.5em{\scshape i\kern-0.25em b}\kern-0.8em\TeX}}}

\usepackage{xspace}
\usepackage{todonotes}
\usepackage[normalem]{ulem}
\usepackage{glossaries}
\usepackage{paralist}

\def\figref#1{Fig.~\ref{#1}}
\newcommand{\babsimhospital}{\textsc{BaBSim.Hospital}\xspace} 
\newacronym{CI}{CI}{Continuous Integration}
\newacronym{CICD}{CI/CD}{Continuous Integration / Continuous Deployment}
\newacronym{CD}{CD}{Continuous Deployment}
\newacronym{DES}{DES}{Discrete Event Simulation}
\newacronym{DIVI}{DIVI}{Deutsche interdisziplinäre Vereinigung für Intensiv- und Notfallmedizin}
\newacronym{ECiP}{ECiP}{Evolutionary Computation in Practice}
\newacronym{EGO}{EGO}{Efficient Global Optimization}
\newacronym{ETL}{ETL}{Extract, Transform, and Load}
\newacronym{ICU}{ICU}{Intensive Care Unit}
\newacronym{LHD}{LHD}{Latin Hypercube Design}
\newacronym{RKI}{RKI}{Robert Koch Institute}
\newacronym{RMSE}{RMSE}{Root Mean Squared Error}
\newacronym{RSM}{RSM}{Response surface methodology}
\newacronym{simmer}{simmer}{``Discrete-Event Simulation for R''}
\newacronym{SHERPA}{SHERPA}{Simultaneous Hybrid Exploration that is Robust, Progressive and Adaptive}
\newacronym{SMBO}{SMBO}{Surrogate Model-Based Optimization}
\newacronym{SMBSA}{SMBSA}{Surrogate Model-Based Sensitivity Analysis}
\newacronym{SPOT}{SPOT}{Sequential Parameter Optimization Toolbox}

\usepackage{color}
\usepackage{fancyvrb}

\DefineVerbatimEnvironment{Highlighting}{Verbatim}{commandchars=\\\{\}}
\usepackage{framed}
\definecolor{shadecolor}{RGB}{248,248,248}

\renewenvironment{description}[0]{\begin{compactdesc}}{\end{compactdesc}}
\begin{document}
%The new ACM Consolidated TeX template Version 1.3 and above automatically creates and positions these text blocks for you based on the code snippet 
%which is system-generated based on your rights management choice and this particular conference.
%Please copy and paste \setcopyright{rightsretained} before \begin{document} and please copy and paste the following code snippet into your TeX file between \begin{document} and \maketitle, either after or before CCS codes.

%\copyrightyear{2021} 
%\acmYear{2021} 
%\setcopyright{rightsretained} 
%\acmConference[GECCO '21 Companion]{2021 Genetic and Evolutionary Computation Conference Companion}{July 10--14, 2021}{Lille, France}
%\acmBooktitle{2021 Genetic and Evolutionary Computation Conference Companion (GECCO '21 Companion), July 10--14, 2021, Lille, France}
%\acmConference[]{}{}{}
%\acmDOI{}
%\acmISBN{978-1-4503-8351-6/21/07}

%\title{Please do not edit until noon, Feb 4th}
\title{Resource Planning for Hospitals Under Special Consideration of the COVID-19 Pandemic:\\ Optimization and Sensitivity Analysis}
\titlenote{Extended version of work presented at GECCO'21.}

\author{Thomas Bartz-Beielstein}
\email{thomas.bartz-beielstein@th-koeln.de}
\affiliation{%
  \institution{TH Köln}
  \city{Cologne}
  \country{Germany}
}

\author{Marcel Dröscher}
\email{marcel_ulrich.droescher@smail.th-koeln.de}
\affiliation{%
  \institution{TH Köln}
  \city{Cologne}
  \country{Germany}
}

\author{Alpar Gür}
\email{alpar.guer@smail.th-koeln.de}
\affiliation{%
  \institution{TH Köln}
  \city{Cologne}
  \country{Germany}
}

\author{Alexander Hinterleitner}
\email{Alexander.Hinterleitner@th-koeln.de}
\affiliation{%
  \institution{TH Köln}
  \city{Cologne}
  \country{Germany}
}

\author{Olaf Mersmann}
\email{olaf.mersmann@th-koeln.de}
\affiliation{%
  \institution{TH Köln}
  \city{Cologne}
  \country{Germany}
}

\author{Dessislava Peeva}
\email{dessislava_todorova.peeva@smail.th-koeln.de}
\affiliation{%
  \institution{TH Köln}
  \city{Cologne}
  \country{Germany}
}

\author{Lennard Reese}
\email{lennard.reese@smail.th-koeln.de}
\affiliation{%
  \institution{TH Köln}
  \city{Cologne}
  \country{Germany}
}

\author{Nicolas Rehbach}
\email{nicolas_alexander.rehbach@smail.th-koeln.de}
\affiliation{%
  \institution{TH Köln}
  \city{Cologne}
  \country{Germany}
}

\author{Frederik Rehbach}
\email{frederik.rehbach@th-koeln.de}
\affiliation{%
  \institution{TH Köln}
  \city{Cologne}
  \country{Germany}
}

\author{Amrita Sen}
\email{amrita.sen@gmx.de}
\affiliation{%
  \institution{TH Köln}
  \city{Cologne}
  \country{Germany}
}

\author{Aleksandr Subbotin}
\email{aleksandr.subbotin@smail.th-koeln.de}
\affiliation{%
  \institution{TH Köln}
  \city{Cologne}
  \country{Germany}
}

\author{Martin Zaefferer}
\email{Martin.Zaefferer@th-koeln.de}
\affiliation{%
  \institution{TH Köln}
  \city{Cologne}
  \country{Germany}
}
\authornote{Authors are listed alphabetically}

\renewcommand{\shortauthors}{Bartz-Beielstein et al.}

\begin{abstract}
    Crises like the COVID-19 pandemic pose a serious challenge to health-care institutions. They need to plan the resources required for handling the increased load, for instance, hospital beds and ventilators. To support the resource planning of local health authorities from the Cologne region, \babsimhospital, a tool for capacity planning based on discrete event simulation,  was created. The predictive quality of the simulation is determined by 29 parameters. Reasonable default values of these parameters were obtained in detailed discussions with medical professionals.  We aim to investigate and optimize these parameters to improve \babsimhospital. First approaches with "out-of-the-box" optimization algorithms failed. Implementing a surrogate-based optimization approach generated useful results in a reasonable time. To understand the behavior of the algorithm and to get valuable insights into the fitness landscape, an in-depth sensitivity analysis was performed.  The sensitivity analysis is crucial for the optimization process because it allows focusing the optimization on the most important parameters.
    We illustrate how this reduces the problem dimension without compromising the resulting accuracy.
    The presented approach is applicable to many other real-world problems, e.g., the development of new elevator systems to cover the last mile or simulation of student flow in academic study periods.
\end{abstract}
%%https://www.overleaf.com/project/60055feddc5ad9382a02ae29
%%
%% The code below is generated by the tool at http://dl.acm.org/ccs.cfm.
%% Please copy and paste the code instead of the example below.
%%


%%
%% Keywords. The author(s) should pick words that accurately describe
%% the work being presented. Separate the keywords with commas.

%%
%% This command processes the author and affiliation and title
%% information and builds the first part of the formatted document.
\maketitle

\pagestyle{plain}% removes running headers

\section{Introduction}\label{sec:intro}
Our initiative is motivated by the challenges that health care institutions face in the current COVID-19 pandemic. 
Planning the demand and availability for specific resources, such as intensive care beds, ventilators, and staff resources, is crucial.
Policies and decisions made by hospital management professionals as well as political officials need to be well informed to be effective.

This article reports the experiences that were collected over the last 12 months and provides answers to the following questions:
\begin{compactenum}[(Q-1)]
\item How to automate data collection and curation? 
\item How to select a suitable simulation model?
\item How to find an optimization algorithm that is able to solve noisy, dynamic, high-dimensional real-world problems?
\item How to integrate domain knowledge and how to analyze simulation output?
\end{compactenum}
In the following, a holistic approach that demonstrates how tools from evolutionary optimization, simulation, sensitivity analysis, and machine learning can be combined to predict and understand demanding resource allocation problems is presented.
We illustrate how the pieces can be put together in a complex software project, i.e., we consider the collection of noisy, dynamic, and heterogeneous data, data preprocessing, surrogate models to accelerate simulation, the optimization of the model parameters, and a parameter sensitivity analysis.

Simulation models are valuable tools for resource usage estimation and capacity planning.
They can either be implemented top down, e.g., using time series approaches~\cite{Hynd08b} or bottom up, e.g., using \gls{DES}~\cite{Bank01a}.
Benefits of \gls{DES} are manifold and range from providing insights into the process’s risk and efficiency when estimating the effects of alternating configurations of the system.
It helps to gain insight into consequences of redesign strategies.
\gls{DES} has been successfully applied to problems that model customers arriving at a bank,   
products being manipulated in a supply chain, and the performance of configurations of a telecommunications network~\cite{Bank01a}.

This bottom-up approach, i.e., a \gls{DES}, is used to model the hospital resource planning problem.
\babsimhospital simulates the path of many thousands or possibly even millions of patient trajectories through hospitals.
This simulation requires considerable computational resources.
Therefore, a very efficient simulator is required, since only a limited number of simulations can be performed in a reasonable time frame.
We have chosen \gls{simmer}, a \gls{DES} package which enables high-level process-oriented modeling~\citep{Ucar19a}. 
The code required for running the simulations is published as an open-source R-Package
%~\citep{Anon21b}.
~\cite{bart20rArxiv, bart20t}.
%TODO: anonymized. Replace for cam-ready.

\gls{simmer} is based on the concept of a trajectory: a common path in the simulation model for entities of the same type.
Trajectories consist of a list of standardized actions, which define the life cycle of equivalent processes. 
It takes available hospital data into account and offers hospitals and policymakers a means to simulate the progression of the pandemic in terms of available and occupied hospital resources and capacity. 
The modeling approach is inspired by~\citet{Lawt19a} and is enhanced by a  \gls{SMBO} approach~\citep{Forr08a}, i.e., our system 
combines several powerful approaches: 
\begin{description}
    \item[Discrete event simulation:] the 'simmer' R-package is used to generate a simulation with 29 parameters with default values, established in cooperation with medical professionals~\citep{Ucar19a}. These parameters are essential for the accuracy of the simulation and require careful optimization. Although domain knowledge, i.e., from medical professionals, provides valuable information to perform realistic simulations, further fine-tuning is required.
    \item [Model-based optimization:] the  \gls{SPOT} R-package is used to perform  \gls{SMBO} to identify the best values for the 29 parameters in a fast and accurate manner, which results in an optimization-via-simulation approach~\cite{Fu94a}.
\end{description}
However, the relatively large number of parameters limits the quality of the optimization process. 
In addition to the improved validity and performance of the simulation model, the \gls{SMBO} approach generates an important benefit: 
\begin{description}
\item[Sensitivity analysis:] results from the  \gls{SMBO}, i.e., models and data, can be directly used to perform a sensitivity analysis~\citep{Salt08b}.
Because our approach relies heavily on these surrogate models, it will be referred to as \gls{SMBSA}.
\end{description}
In that context, the data collected during optimization is analyzed using a Kriging model, a linear model, and a Random Forest model to identify the important parameters as well as parameters that exercise almost no impact on the modeling process. 
We illustrate a technique for evaluating parameter importance in \babsimhospital.
These results may also outline a way to reduce problem dimensions without compromising accuracy.

The rest of this paper is structured as follows: 
Section~\ref{sec:data} discusses the available data and its preparation,
Section~\ref{sec:sim} introduces the \babsimhospital simulator and 
Section~\ref{sec:optim} describes the corresponding optimization problem. 
Section~\ref{sec:sa} presents sensitivity analysis.
The  experimental setup is presented in Section~\ref{sec:experiments}.
This section also discusses reproducibility and technical issues.
Results from the sensitivity analysis are described in Section~\ref{sec:results}.
Finally, the results of this project are discussed in Section~\ref{sec:discussion}.

\section{Automated Data Collection and Curation}\label{sec:data}
\gls{ETL} processes integrate data from various sources into complex collections. 
A typical problem is ``that each data source has its distinct set of characteristics 
that need to be managed in order to effectively extract data for the \gls{ETL} process''~\citep{ELSAPPAGH201191}. 
After the successful extraction of data, the next step is to transform it. 
This step includes several approaches to gain accurate data which is correct, complete, consistent, and unambiguous.  
The final step consists of loading the processed data into a data collection of choice accessible for the data analyst for further use.
Especially in terms of the COVID-19 pandemic, it is important to integrate and process the vast amount of constantly growing data. 
Therefore, \babsimhospital implements an \gls{ETL} process to analyze the data from the \gls{RKI}, \url{https://www.rki.de}, as well as the \gls{DIVI}, \url{https://www.divi.de}).
The associated data sets contain anonymous information about every recorded case in Germany. The \gls{RKI} data set contains 
780,065 observations of 18 variables such as age, gender, data of infection, etc., which were updated daily and are automatically integrated into \babsimhospital. 

Information concerning \gls{ICU} in Germany can be retrieved from the \gls{DIVI}.
\gls{DIVI} provides an API and a daily report. The API does not provide all data the daily report consists of and therefore is not a viable option for this project.
Web scraping was implemented as a reliable solution to retrieve the most current daily report from \gls{DIVI}.

\section{The Simulator}\label{sec:sim}

\begin{figure}
    \centering
    \includegraphics[width=0.95\linewidth]{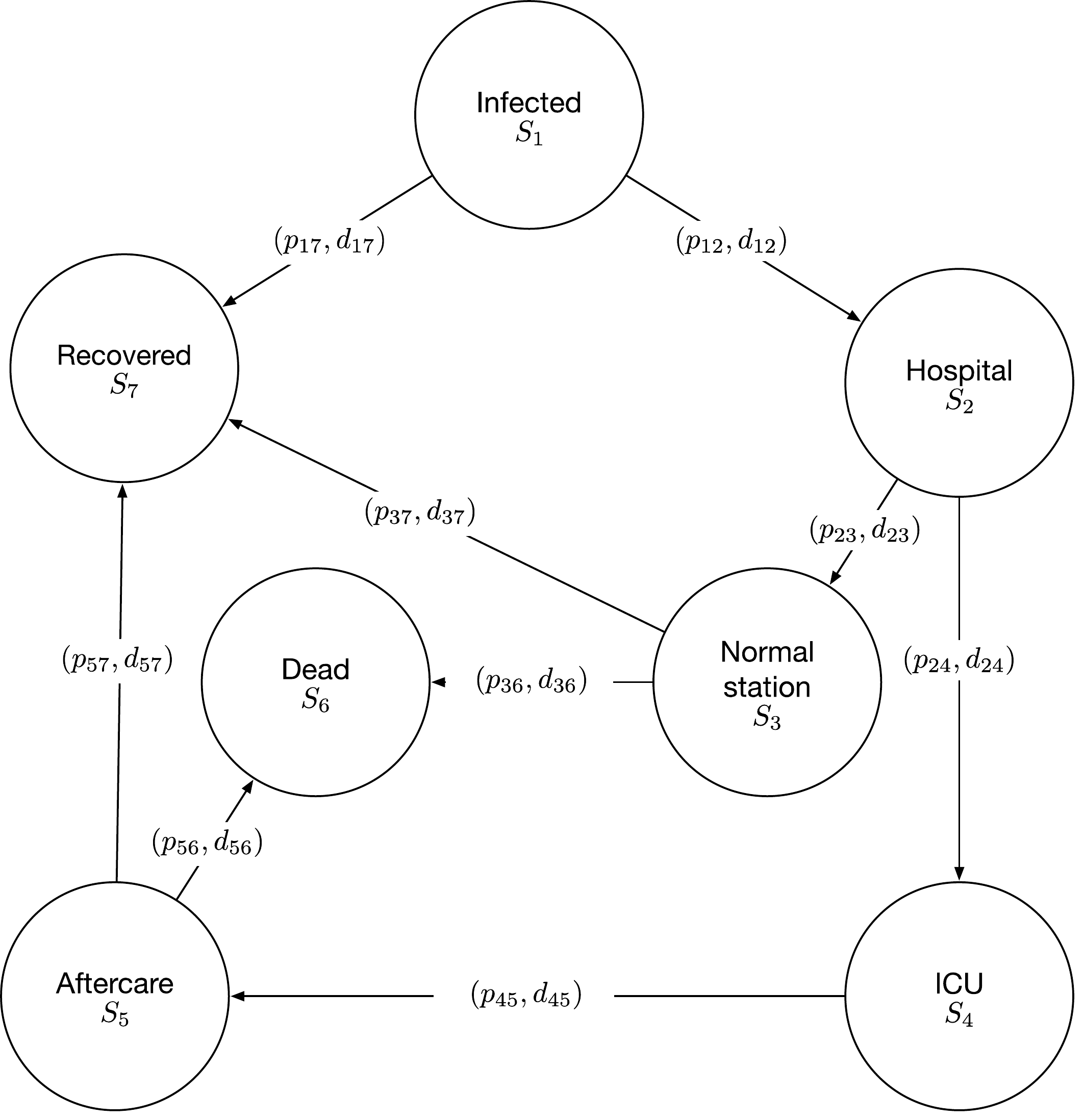}
    \caption{Simplified model of patient flows in a hospital. Nodes represent states ($S_i$). Edges represent state changes with associated probabilities ($p_{ij}$) and durations ($d_{ij})$.
    Probabilities and durations at time step $t$ of the optimization will be collectively referred to as \emph{model parameters} $\vec{x}_t$.}
\label{fig:hospital}
\end{figure}

\babsimhospital simulates the typical paths that COVID-19 infected patients follow during their hospital stays. 
The \gls{DES} processes every single recorded infection until the patients' recovery or death. 
Patients follow a trajectory, i.e., they move with a probability $p_{ij}$ from state $S_i$ to state $S_j$  after a transition-specific duration $d_{ij}$.
A graph can be used to model this behavior (\figref{fig:hospital}).
For example, an infected patient (state $S_1$) goes to the hospital (state $S_2$) with probability $p_{12}$ after $d_{12}$ days. With probability $p_{17}$, she recovers (state $S_7$) after $d_{17}$ days. 
The probabilities of outgoing nodes sum to 1, e.g., $p_{17} = 1 - p_{12}$.
The modeling process includes four types of parameters: 
\begin{description}
\item[transition probabilities,] e.g., the probability that an infected individual has to go to the hospital,
\item[durations,] e.g., the time span until an infected individual goes to the hospital (in days), and
\item[distribution properties,] e.g., truncated and translated gamma distribution,
\item[risk factors] depending on demographic groups, e.g., age, gender. 
\end{description}

The "risk" attribute is an important factor for the duration and severity of a COVID-19 infection. 
Statistics show that the risk depends on the age and gender of the infected person. 
In our simulation, this risk is stored as an attribute of the patients. 
Every new patient is assigned a unique risk. 
We use an exponential function to model the relationship between the age of the person and their risk.
That means that older patients have a much higher probability of requiring an intensive bed or an intensive bed with ventilation than younger patients. 
In addition, men have a 50 percent higher risk than women. 

Proper tuning of these parameters is essential to obtain accurate predictions based on up-to-date and local data. 
The time-dependent changes require a frequent refitting of the model parameters to the current situation.
Thus, a daily parameter tuning procedure is run for each German region in order to provide an accurate prediction with \babsimhospital. 
An initial estimate for each of the given parameters was specified in cooperation with medical professionals. 
For example, the rate of successful treatments in Germany drastically changed between the first and the second wave of COVID-19 infections. 
Also, political decisions on national and local level can affect the situation significantly. 
While reducing the access to nursing homes might reduce infections in the high risk parts of the population, opening schools might cause many infections in the younger parts of the population. The optimization problem can be stated as follows:
the \babsimhospital simulator requires two input parameters:
\begin{compactenum}
\item $\vec{x}_t$, the model parameters
\item $\vec{u}_t$, the number of infections.
\end{compactenum}
Based on these two inputs, \babsimhospital estimates the required resources---in our case, the beds, \gls{ICU} beds,  and \gls{ICU} beds with ventilators.
The simulation output, i.e, the required resources on each day $t$ will be denoted as $\hat{\vec{y}}_t$, i.e., 
\begin{equation}\label{eq:haty}
\hat{y}_t = \left( \hat{R}_{\text{bed}}(t),   \hat{R}_{\text{icu}}(t), \hat{R}_{\text{vent}}(t) \right)
\end{equation}
Although the tool was designed to assist the German local authorities, it was already adapted and used for resource planing in hospitals in other countries.

\section{Optimization}\label{sec:optim}
After the optimization-via-simulation problem is defined, this section describes the current state in optimization and discusses the selection of a suitable optimizer. 

\subsection{Optimization Problem}
Based on the simulation results, optimization runs can be performed to improve parameter settings proposed by the experts. 
The \gls{RMSE} as shown in Eq.~\ref{eq:rmse}, is used to measure the error of the simulator.
We formulate the \emph{minimization} problem:
\begin{equation}\label{eq:rmse}
  \min 
  \sum_{k\in\{\text{bed},\text{icu},\text{vent}\}}
    w_k  \sqrt{\frac{1}{T} \sum_{t=1}^T \left(R_k(t) - \hat{R}_k(t)\right)^2}
\end{equation}
Here, $T$ denotes the number of days simulated and $k$ the three different bed categories.
Since the different bed types are not equally important a weighted average of the \gls{RMSE} for each bed category is used as the final error measure.
A detailed description can be found in~\citep{Bart20j}.

The extensive amount of data that the tool has to process combined with the high dimensionality of the problem, and the required accuracy make simplifying the modeling process to improve performance a big challenge.
The limited time available for each optimization run requires the use of efficient algorithms.
To resolve that problem, a number of different optimization-via-simulation techniques can be used. 
These will be covered in the following.

\subsection{Standard Optimizers}
First, the applicability of standard optimization algorithms, e.g., BOBYQA~\citep{Powe09a}, CMA-ES~\citep{Hans06a}, Simulated Annealing~\citep{vanL87a}, was tested. Pre-experimental results revealed that these optimizers were not able to find an improved parameter configurations under the hard time constraints posed by this problem. 

\subsection{Parallelizing and  Combining Global with Local Optimizers}\label{sec:comb}
\citet{Good18b} presented during last years' GECCO \gls{ECiP} track approaches that implement combinations of global and local search methods with a special focus on black-box optimizers.
He discussed optimization strategies that combine different optimization methods. 
For example, the commercial software HEEDS contains a unique search strategy called \gls{SHERPA}~\citep{Good08a}.
During a single search, \gls{SHERPA} uses multiple search methods in parallel.
This approach takes advantage of the strengths of each method, and reduces a methods’ participation in the search if/when it is determined to be ineffective.
However, because \gls{SHERPA} is a commercial software, it was not applicable to our problem.

\subsection{Massively Parallel Single-Iteration Optimizers}\label{sec:parallel}
During the last years the availability of parallel computation increased significantly. 
Even personal computers are equipped with multi-core processors and allow simple parallel optimizations. Recent approaches take parallelization to the extreme: why not use several thousands, or millions of processors in parallel for a so-called \emph{one-shot} optimization run?
In one-shot optimization, aka single-iteration evolution or fully-parallel optimization, the user selects a population, evaluates it, and has to base all future decisions only on the quality of these points.
In a recent work, \citet{Cauw20a} analyzed the setting in which an optimal solution is chosen at random from a Gaussian distribution.
They could prove that, unlike one might expect, it is better to sample only one (namely, the center of the distribution) rather than sampling $n$ times from the same Gaussian distribution.

Fully-parallel optimization was also proposed for hyperparameter search~\citep{Cauw20a}.
The authors  demonstrated that it can be be more effective than random and grid search and that performance improvement can be obtained in a wide range of expensive artificial intelligence tasks.
Furthermore, \citet{Rena20a} discuss the sensitivity of the fitness landscape analysis to fully-parallel optimization sampling strategies, which is also of great relevance in this context, because the sampling strategy is crucial in fully-parallel optimization.
The fully-parallel optimization approach is used in industry by companies like Facebook, that have large server farms at their disposal.
It is considered as the only possible solution if optimization has to be done under very hard time constraints. 
However, this approach was not applicable for our optimization, because we only have a few hundred CPU cores available.

\subsection{Response Surface Methodology and Surrogate Model-based Optimizers}\label{sec:rsm}
\gls{RSM} is a collection of statistical and mathematical tools useful for developing, improving, and optimizing processes~\citep{Myers2016}. 
Applications historically come from industry and manufacturing, 
However, in many settings, their intended application is too local. Moreover, \gls{RSM} is ``too hands-on''.
Because computational power is available in every real-world optimization scenario, 
it might be useful to automate, i.e., to remove humans from the loop and set the computer running on the optimization in order to maximize computing throughput, or minimize idle time. 
Therefore, new response surface methodologies were developed in the last decades~\citep{Gram20a}.
 
 One prominent representative in very budget-restricted environments is \gls{SMBO}~\citep{Jin11a}.
In \gls{SMBO}, a data-driven surrogate is fitted to only a few initial sample points on the expensive to evaluate objective function. 
The fitted model is (compared to the real objective function) computationally cheap to evaluate.
Thus, promising new candidate solutions can be proposed by running an extensive optimization on the model.
The search on the model is guided by a so-called infill criterion or often also acquisition function.
The function assigns a quantitative quality, usually based on the models prediction and uncertainty, to a given candidate.
The best candidate proposed by the model is evaluated on the expensive objective function and a new model is fitted including the new data point.
This process is iterated until the budget of expensive evaluations is depleted or some other stopping criterion is met. More in-depth explanations of model-based optimization and its applications can be found in~\citet{Quei05a, EGOB02, Jin19a}.

Only \gls{SMBO} approaches produced good results. 
Therefore, the \gls{SPOT} algorithm was chosen~\citep{Bart17parxiv}.
To accelerate the \gls{SPOT}, simulations were run in parallel.
Besides robust and relatively fast optimizations, \gls{SPOT} provides an additional advantage: surrogate models used by the optimizer can be used (``recycled'') for a sensitivity analysis. 

\section{Model-Based Sensitivity Analysis}\label{sec:sa}
After describing the selection of the simulation model and the optimizer, we discuss 
methods to analyze results from the optimization-via-simulation approach. 
As per \citet{Ustinov15}, at the time of obtaining the best possible outcome from a model numerically, the analytical relation between the model input parameters and
corresponding outcomes are not observed closely. This oversight results in the model potentially containing parameters that are not required. This is where sensitivity analysis provides much needed clarity.
Sensitivity analysis can be used to determine the robustness of model predictions to parameter values and to discover parameters with a high impact~\citep{Rosa18}. There has been extensive research on sensitivity analysis as applied to modeling in various fields. Some of these approaches are highly applicable to analyzing parameter importance for high dimensional models like \babsimhospital. 

\citet{Ahmed20} describe varying parameter values by very small increments to measure and compare the impact of the change on a model for forecasting of COVID-19 using logistic growth models. The study involves optimization algorithms such as Nelder-Mead, Levenberg-Marquardt, and Trust-Region-Reflective~\citep{NelderM65}. The authors use their technique to gather information across these algorithms without the intention of simplifying the modeling process. 
The technique has also been suggested by~\citet{Kleijnen97}, who defines sensitivity analysis as what-if analysis.
There, it is observed how the response of model changes, for a change in input parameters. 

Part of our analysis builds on the ideas of \citet{Ahmed20} and \citet{Kleijnen97}, more specifically we use a similar approach to measure importance by varying parameter values. Subsequently, the conclusions from this and other analytical techniques are combined and used to provide an objective preselection of parameters to be tested for removal from the optimization process.

The approach presented in this paper can be referred to as \gls{SMBSA}, because it uses \gls{SMBO} data and information to perform the sensitivity analysis.
We use the following models in \gls{SMBSA}:
\begin{description}
\item[Linear regression models] were included, because they are well established and considered standard in many classical sensitivity analysis scenarios~\citep{Klei04e}.
\item[Kriging] is a frequently used surrogate in  \gls{SMBO}. It is also referred to as Gaussian process regression~\citep{Scho97a}.
In Kriging, a distance-based correlation structure is determined with the observed data~\citep{Forr08a}. 
Depending on the modeled function, Kriging can deliver accurate predictions even if only a few data points are available.
Additionally, it is often favored for being able to estimate its own prediction uncertainty. 
\item[Random Forests]  were initially introduced by~\citet{breiman2001random}.
They use an ensemble of tree predictors where each tree is fitted to a random subset of the data.
Random Forests are well-established as they are very robust to different variable classes.
They can be applied to classification as well as prediction tasks with continuous, mixed, or discrete data.
Among others, random forests gained publicity in being used for regression and classification through~\citet{liaw2002classification}. 
\end{description}

\section{Experimental Setup}\label{sec:experiments}

\subsection{Steps Performed in this Analysis}
To ensure a detailed and comprehensive understanding of the parameters, the following procedures were employed. 

\subsubsection*{Step 1: Domain Analysis}
A domain analysis provided for a qualitative comparison and a preliminary classification of high and low importance parameters, based on their expected real-life impact. Each parameter was discussed and its relation to the hospital resource allocation was evaluated. The purpose was not to create a comprehensive ranking of the 29 parameters but rather to identify parameters with a significantly higher or lower expected impact than all others, e.g., parameters that would not affect the hospital admissions at all.

\subsubsection*{Step 2: Experimental Exploration}
A set of \glspl{LHD} \cite{McKa79a} were generated to create evenly spread out sample points over the entire feasible search space.
By doing so, we tried to ensure that we obtain a globally valid estimate of parameter importance that is less affected by local anomalies.
The obtained data were fit Kriging-, Linear-, and Random Forest models for further analysis.

\subsubsection*{Step 3: Importance Index}
The \emph{importance index}\/ of the $i$-th parameter was calculated as $P_i = \sum_{k=1}^{n} O_i$,
where $O_i$ is the position of the $i$-th parameter $x_i$ in the sequence of parameters sorted from most important to least important, i.e., descending, 
and $n$ is the number of runs (repeats) of the modeling technique. 
For our experiment, $i \in \{ 1,2, \ldots, 29\}$, with $n = 20$ and $\max(O_i)= 29$.
The \emph{normalized importance index}\/ was determined  as 
\begin{equation}\label{eq:nimp}
    P^{\ast}_{i} = \frac{1}{n \times \max(O_i)}\times\sum_{k=1}^{n} O_i.
\end{equation} 
The normalized values can be seen on Figure \ref{fig:importanceByType}.
Although the models have their own numerical estimates for importance, this data was not taken into account but rather only the ranking of the importance of the parameters. Ignoring this potentially valuable information was crucial to enable us to perform cross-model comparisons.
Due to the large number of parameters, the linear regression model used main effects only. 

\subsubsection*{Step 4: Parameter Impact on the Simulation Error}
Apart from the estimation of the parameter importance through different  modeling techniques, another test was performed similar to the approach presented in~\cite{Ahmed20}. 
The sets of parameters from the \gls{SPOT} initial design were used.
For one parameter set, $n$ simulation repeats were performed and the mean simulation error $E_{\mu}$ was calculated.
Then, each parameter, one by one, was slightly changed and a new mean simulation error $E_{\mu_i}$ was determined for the same number of simulations.  
Finally, we calculated the differences in errors for each parameter 
\begin{equation}\label{eq:delta}
\Delta E_{x_i} =  \frac{|E_{\mu}-E_{\mu_i}|}{E_{\mu}}\times 100 \%.
\end{equation}
The index $i$ represents the changed parameter number. 
This experiment was repeated for $k=33$ available parameter configurations. 
Results are shown in Figure \ref{fig:errorChangebarplot}.

\subsubsection*{Step 5: Parameter Plots}
Selected parameters were compared using parameter plots
that illustrate the impact of each parameter on the model error. 
These experiments aim to bring further clarity on the parameter interaction and importance.
The analysis based on the above described experiments is discussed in more detail in the next section. 
Its purpose is to enable a reliable proposal for most and least important parameters.

\subsubsection*{Step 6: Optimization Tests}\label{sec:optimTests}
As a final step, after sufficient information is gathered on the parameter importance, 
an attempt to use the information in order to gain optimization efficiency can be made. 
Parameters which only marginally change the outcome of a simulation could potentially be excluded from the optimization. 
Then, the very limited budget of the optimizer can be spent more efficiently on the remaining more important parameters. 

A set of experiments was executed to investigate this question.
The SPOT-Direct optimizer was run for 30 repeats in an attempt to tune the simulations parameters.
Then, one parameter after another was excluded iteratively from the optimization.
The excluded parameters are fixed to the mean between their bounds and fed into the simulation together with the other parameters.
Therefore, they are still part of the simulation, but are not being optimized.
This part provides a proof of concept by observing the actual optimization performance with optimization limited to only selected parameters. 

\subsection{Reproducibility, CI/CD, and Testing}\label{sec:reproduce}

The number of corona virus infections strongly varies over time and also for each tested region. 
To ensure reproducibility of our experiments, it is therefore necessary to select and fix the specific data that was chosen.
As this project was initiated together with local health officials, the data for the German city of Cologne was chosen for the in-depth analysis.
Each simulation is initially fed with four weeks worth of cases as a \emph{warm-up}. 
After that, the actual evaluation time-frame starts in which the quality of the simulation is measured. 
Field data is only required for this  second phase.
The infection data was therefore chosen from 2020/10/12 to 2021/01/14.
The field data from 2020/11/09 to 2021/01/14.
While we had to choose a specific region and period for the experiments described here, they can easily be adapted s for other regions or time periods.
Furthermore, as this project aims to make the tools to create such analysis available to the public, all software required to perform the experiments and simulations is available online: 
\url{https://cran.r-project.org/web/packages/babsim.hospital/index.html}.

The data that \babsimhospital uses is constantly growing and changing.
Similarly, features of the \babsimhospital software are added or changed continuously.
Therefore, it is crucial to maintain a \gls{CI} process. 
It has six stages: the \texttt{install}, \texttt{test}, and \texttt{deploy} stages integrate new features and code changes reliably.
The \texttt{optimparas} stage runs the parameter optimization process for each region (e.g., country / state / district).
The \texttt{preload} stage precomputes results so users can analyze the results interactively. 
Lastly, the \texttt{updatedata} stage is responsible for updating the production data and performing the \gls{ETL} process described in Section~\ref{sec:data}.
The \gls{CICD} process is fully-automated, using a GitLab server.

To automatically ensure that the simulation runs without errors, we developed a series of tests using the R package \textsc{testthat}~\citep{Wick11a}. 
Different possible paths of the patients were simulated deterministically as well as stochastically. 
Resulting values were compared to the expected resource usage and durations.

\section{Results}\label{sec:results}

\subsection{Domain Analysis}
As described in more detail in \emph{Step 1. Domain Analysis} of Section \ref{sec:experiments}, a domain analysis was performed on the 29 parameters used by the simulation
~\cite{Bartz20}. 
%TODO: anonymized. replace for cam-ready.
Three parameters were established as potentially holding the highest level of importance and three as being potentially irrelevant.

\begin{compactitem} 
\item[$x_{14}$:] The parameter \texttt{FactorPatientsInfectedToHospital} is one of the most important parameters in this simulation. It reflects the percentage of infected people who have to be admitted to the hospital. This affects the general number of infected people that come into the hospital system and is a measure of the severity of the virus.
\item[$x_{13}$:] The parameter \texttt{GammaShapeParameter} determines the distribution and, as such, can be expected to have a profound effect on the model.
\item[$x_{26}$:] The parameter \texttt{ RiskFactorB} may also hold importance as it is an influential part of the modeling of the risk as a function of the age. Similar to $x_{14}$, this parameter would influence the overall outcome for all patients and would hold explanatory value for understanding the variations in other parameters.
\end{compactitem}
Parameters that are expected to have the least importance are $x_{2}$, \texttt{AmntDaysNormalToHealthy}, $x_{5}$, \texttt{AmntDaysNormalToDeath}, and $x_{19}$ \texttt{FactorPatientsNormalToDeath}. All of the above do not involve any hospital stay and should not have a significant impact on the model. 
Other parameters also have a meaningful impact on the model but are not considered highly important. 
For example, although there is no difference in the COVID-19 infection rates of men and women, gender $x_{27}$, \texttt{RiskMale} can be considered an important factor~\citep{Peckham20}. However, due to the higher life expectancy of women, i.e., more old female patients, and the gender imbalance of cardiovascular disease, $x_{27}$ \texttt{RiskMale} is not expected to be a dominating parameter in the model.

\subsection{Empirical Analysis}
\begin{figure}
  \centering
  \includegraphics{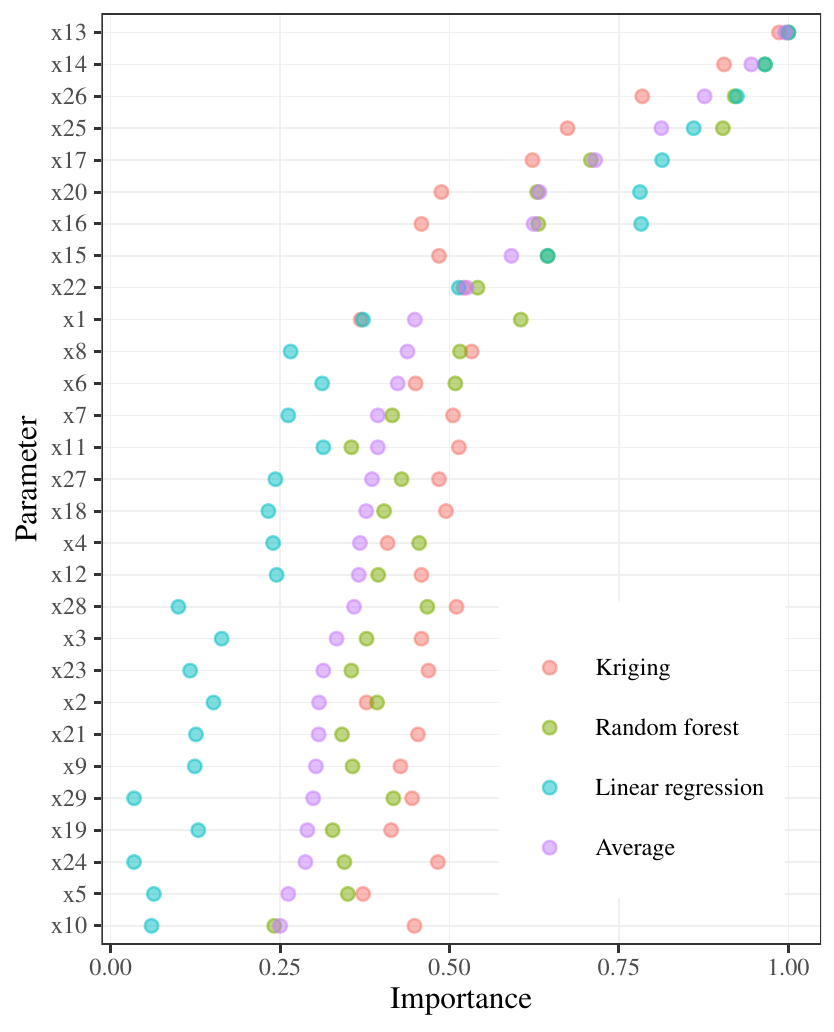}
  \caption{Importance Score Per Modeling Type. The mean importance score of the $n=20$ model runs for each type is used to calculate the normalized importance index $P^{\ast}_i$ of the $i$-th parameter $x_i$ as defined in Equation~\ref{eq:nimp}. The importance index values are displayed for each model type and the parameters are sorted by their ranking.}
\label{fig:importanceByType}
\end{figure}

As described in \emph{Step 3. Calculating an Importance Index} from Section \ref{sec:experiments}, 
the normalized importance index $P^{\ast}_i$ was generated to identify a ranking of parameters with respect to Kriging-, linear-, and Random Forest models.
To compare the results, Figure~\ref{fig:importanceByType} reports the normalized importance $P_{i}$ as assigned by each of the models.

Based on the importance index, the five most important parameters (\emph{$x_{13}$, $x_{14}$, $x_{26}$, $x_{25}$ and $x_{17}$}) are the same for the three models. However, the bottom five parameters differ considerably. Only \emph{$x_{5}$} is identified to be among the five least important  parameters for all models. 
This indicates that identifying the least important parameters is not straightforward.

\subsection{Parameter Impact on the Simulation Error}

\begin{figure}
    \centering
    \includegraphics{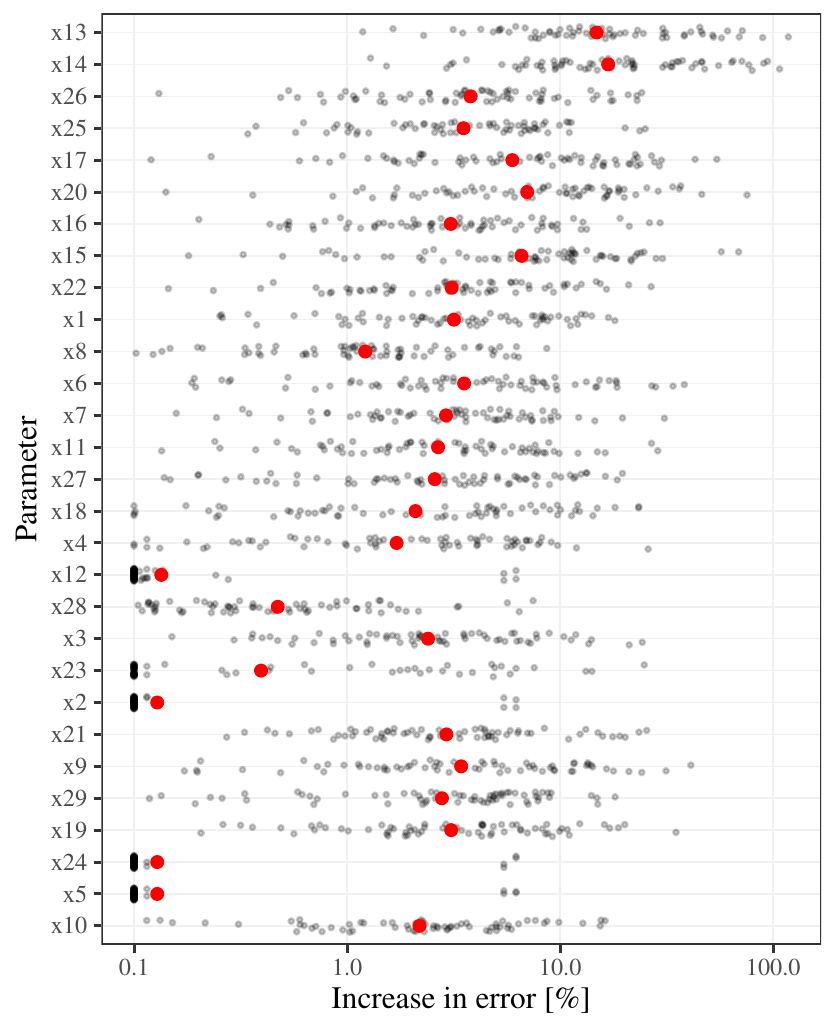}
    \caption{%
    Change in error $\Delta E_{x_i}$ as defined in Equation~\ref{eq:delta} when changing one parameter. 
    Mean change in error (\emph{red}\/ dot) is calculated by increasing and decreasing one parameter value 
    by 20\% while keeping all other parameters unchanged for $k=33$ initial configurations (\emph{black}\/ points).
    Note that the \% change is shown on a log scale.}
\label{fig:errorChangebarplot}
\end{figure}

As discussed in \emph{Step 4. Parameter Impact on the Simulation Error} of Section \ref{sec:experiments}, the purpose of this experiment is to quantify the impact on the error that is affected by a change in the value of an important parameter versus that of an unimportant one. Figure~\ref{fig:errorChangebarplot} shows the measured error from multiple re-runs of the simulation with changed parameter values. 
The observed results do not fully confirm the assumption that changing more important parameters will have a bigger impact on the error than changing less important parameters. For parameters \emph{$x_{13}$} and \emph{$x_{14}$}, the change is noticeably more than for any other parameter, 
confirming the hypothesis that these are two of the most influential parameters. For \emph{$x_{26}$}, however, the result is comparable to that of parameters with less estimated importance.

The domain Analysis is partially confirmed: The parameters \emph{$x_{2}$} and \emph{$x_{5}$} have negligible impact.
However, \emph{$x_{19}$} diverges from that expectation, displaying an impact similar to that of \emph{$x_{26}$}. 
The overall results offer assurance for some conclusions of the domain analysis and are partially in compliance with the ranking of the normalized importance index. 

\subsection{Parameter Plots}

\begin{figure*}
  \centering
  \includegraphics{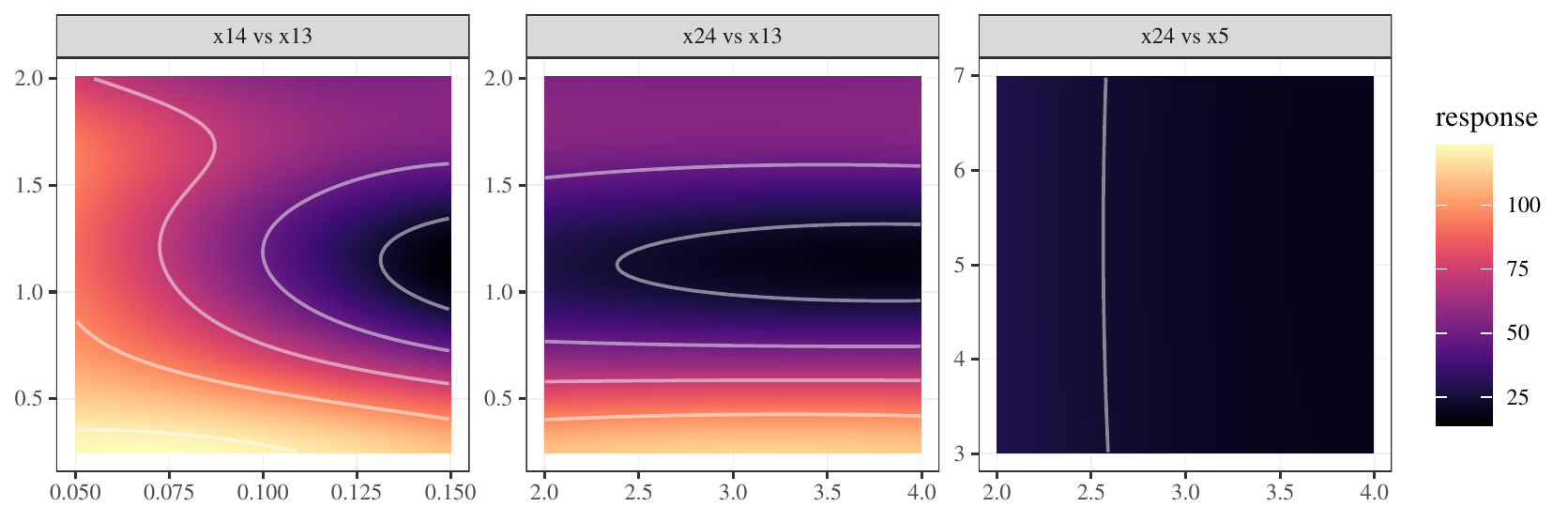}
  \caption{Left: 
  Behavior of two highly important parameters, \emph{$x_{14}$} (x-axis) and \emph{$x_{13}$}. 
  The change in response, which represents the \gls{RMSE} as defined in Equation~\ref{eq:rmse}, is significant and both parameters have a strong impact.
  Center: Behavior of a highly important parameter \emph{$x_{13}$} (x-axis) with respect to a less important one \emph{$x_{24}$}. The change in the response illustrates the importance of \emph{$x_{13}$} for the response (\gls{RMSE}), and the absence of impact of parameter \emph{$x_{24}$}.
  Right: Behavior of two parameters with low importance, \emph{$x_{24}$} (x-axis) and \emph{$x_{5}$}.
  The change in the response is very small.
  }
  \label{fig:model-contours}
\end{figure*}

Parameter plots provide a clear visualization of the range and behavior of the high and low impact parameters and serve as a useful tool to illustrate the parameter dependencies. 
For instance, Figure~\ref{fig:model-contours} displays the range of \emph{$x_{13}$} and the relatively low impact of \emph{$x_{24}$} on the response in the middle panel. 
In the right panel the two parameters with little importance (as per both the domain and the data-driven analysis) exhibit no impact, the magnitude of the response that is affected by them is negligible.
Finally, in the left panel the impact of two highly important parameters on the response is shown.
The parameter plots are a useful functionality of \gls{SPOT}, allowing for the visual representation of the parameter importance and range. They are representative of the model and help confirm the conclusions from the previous analysis.

%- $x_{13}$ is consistently ranked before $x_{14}$. Can we quantify the difference using the parameter plots? Is it possible to narrow the range of values for either parameter?
%- Same for $x_{13}$ and $x_{26}$ and $x_{14}$ to $x_{26}$. $x_{25}$ is another parameter that affects the calculation of risk. Can the parameter plots suggest a range for it that is more valid and perhaps help adjust the default values?

\subsection{Removing Parameters from Optimization}\label{sec:removing}
The presented sensitivity analysis indicates that some parameters have a large impact on the resulting objective function value, while others only show little impact. 
A natural next step is to use this knowledge to increase the optimization efficiency. 

%The results of runs discussed in Section \ref{sec:optimTests} are shown in Figure \ref{fig:OptimizationTests}.
Yet, the strong noise, which results from the simulator, also manifests in our optimization results, making it harder to observe any actual changes in optimizer efficiency when adding or removing single simulation parameters.
Our analysis found that significant changes in optimization quality only occurred when one or more of the five most important parameters were removed from the optimization. 
Removing less important parameters did not show significant changes, or would require more repeats for validation. 
However, repeats of this experiment involve running multiple optimizations on an expensive to evaluate simulation, thus prohibiting in-depth studies.  

Overall the experiments showed that less important parameters might be safely removed from the optimization, therefore, reducing optimizer runtime.
More importantly though the experiments emphasize that the five most important parameters were indeed identified correctly.
Yet, due to the huge overlaying noise, this requires further investigation in future studies.

%\begin{figure}
%    \centering
%    \includegraphics[width=1\linewidth]{variableRemovalBoxPlot.png}
%    \caption{Optimizer performance while tuning the simulation parameters. Lower values indicate better results. From left to right, more and more variables are removed from the optimization and fixed to a constant. The red line fits a smoothed line through all sample points visualizing a general trend in the data.}
%\label{fig:OptimizationTests}
%\end{figure}

\section{Discussion and Outlook}\label{sec:discussion}
The high dimension and computational expense of the \babsimhospital simulator 
poses a challenging optimization task. 
Solving this task for many regions in Germany under very different local circumstances requires 
efficient solutions to cope with the further growing infection numbers and thus also growing simulation run times. 
This article presents a holistic approach to solve this challenging problem. 
It reports experiences from a one-year project with several stakeholders. 
From the Questions posed in Section~\ref{sec:intro} we would like to answer (Q-1) to (Q-3) directly.
\begin{compactenum}[(Q-1)]
\item How to automate data collection and curation?
The system is running fully automatically for several months.
It allows processing  the \gls{RKI} data set, which consists of more than 
750,000 observations of 18 variables, which are updated daily and are automatically integrated into \babsimhospital simulator. 
The \gls{CICD} approach minimizes human interaction, so that simulations and optimizations are
started automatically after the data is downloaded.
\item How to select a suitable simulation model?
The \gls{DES} delivers valid results and enables predictions, which are valuable for capacity planning in hospitals.
The \gls{simmer} software presents a good basis for implementation and was able to handle more than half a million data (infections) under very limited time constraints. 
\item How to find an optimization algorithm that is able to solve noisy, dynamic, high-dimensional real-world problems?
Using "out-of-the-box" optimizers did not work, because the problem is too noisy and high-dimensional.
Applying model-based optimizers to this simulation problem provided good results but also slightly increased the computational burden due to the runtime of the model-based optimizer. 
Using the estimated parameter importances from the sensitivity analysis indicated that running the model-based optimizer with fewer parameters is possible without a significant quality loss. 
Removing parameters from the model-based optimization loop can drastically decrease the optimizers runtime. 
\end{compactenum}
Answering (Q-4) revealed such amounts of interesting material, that we would like to dedicate its own subsection to it:
\subsubsection*{\textbf{(Q-4)} How to integrate domain knowledge and how to analyze simulation output?}
Our approach of determining the most important parameters has provided valuable insights into the simulator.
While there were a few parameters that dominate the simulation quality, many more affect only slight changes in the simulation.
In line with the conclusions from the domain analysis, \emph{$x_{13}$}, \emph{$x_{14}$} and \emph{$x_{26}$} are the parameters with the highest impact on the model.
These findings can support practitioners in their choices.

Determining the least important parameters has not been as straightforward. 
Parameters that should have no impact on the model seem to consistently make a meaningful contribution, albeit not being among the parameters with the strongest explanatory value. 
Comparing the importance based analysis and parameter impact on simulation error, we are able to observe that \emph{$x_{13}$} and \emph{$x_{14}$} are still consistently holding their position as the most important parameters while \emph{$x_{24}$} and \emph{$x_{5}$} are consistently among the bottom five parameters.
There may be multiple reasons for that but the most likely one may stem from the special case of hospital resource planning in the situation of a pandemic that is the subject of this research. 

%$x_{19}$ 
For example, \emph{FactorPatientsNormalToDeath} describes a trajectory of patients that never used an intensive care bed. 
Yet, this factor may be intrinsically related to all parameters that measure for the general level of seriousness of the disease such as \emph{$x_{14}$ FactorPatientsInfectedToHospital}. 
A higher percentage of infected patients admitted to the hospital implies more complications that need hospital care during the course of the disease. 
A disease with an expected high rate of complications might also have a higher \emph{$x_{19}$ FactorPatientsNormalToDeath}. 
The same view would apply to \emph{$x_{2}$	AmntDaysNormalToHealthy} and \emph{$x_{5}$ AmntDaysNormalToDeath}.

%In sum, the power of the modeling process stems from its ability to extract meaning from various sources and combine their explanatory power to create a highly predictive simulation. As much as reducing the number of parameters is a valuable step to simplifying the model, eliminating contributing parameters is counterproductive to the overall goals of the \babsimhospital tool. 

%In the case of the \babsimhospital tool, accuracy is essential as it serves public authorities in a real-life critical situation. Therefore, adjusting the default parameter values is valuable for the public. A reduction of the parameters to simplify and speed the modeling process has mostly technical benefits. There is no contribution to the goal of providing accurate estimates of the future hospital resource needs to the public health authorities, and, therefore, cannot be undertaken based on the obtained results in this study.
\subsubsection*{Conclusion and Outlook}
In conclusion, although results of this study are highly dependent on the medical data used for the modeling and simulation, the general \gls{SMBSA} approach to parameter evaluation for sensitivity analysis can be applied to any model with a large number of parameters.
The possibility of \gls{SMBSA} to reuse the same models that were used in an optimization for sensitivity analysis shows clear advantages by simplifying the overall approach.
Furthermore, the proposed optimization-via-simulation approach is not limited to resource planning problems in hospitals.
We are currently using a very similar approach to develop new elevator systems and to simulate student flow in academic study periods.
Finally, we are performing additional benchmark studies to investigate the behavior of alternative \gls{SMBO} approaches and optimizers on this challenging problem.
The \babsimhospital simulator can be used as attractive simulation and optimization benchmark example. Its source code is open source.

\bibliographystyle{ACM-Reference-Format}
\bibliography{bart21f}
\end{document}